\def\year{2020}\relax
\documentclass[letterpaper]{article} % DO NOT CHANGE THIS
\usepackage{aaai20}  % DO NOT CHANGE THIS
\usepackage{times}  % DO NOT CHANGE THIS
\usepackage{helvet} % DO NOT CHANGE THIS
\usepackage{courier}  % DO NOT CHANGE THIS
\usepackage[hyphens]{url}  % DO NOT CHANGE THIS
\usepackage{graphicx} % DO NOT CHANGE THIS
\usepackage{graphicx}  % DO NOT CHANGE THIS
\usepackage{booktabs}       % professional-quality tables
\usepackage{graphicx}
\usepackage{amsmath,amssymb} % define this before the line numbering.
\usepackage{makecell}
\usepackage{color}
\usepackage{enumitem}
\usepackage{multirow}
\usepackage{xcolor}
\usepackage{romannum}
\usepackage{rotating}
\usepackage{hhline}
\AtBeginDocument{\pagenumbering{arabic}}

\DeclareMathOperator*{\argmax}{arg\!\max}

\newcommand{\ie}{\textit{i}.\textit{e}., }
\newcommand{\eg}{\textit{e}.\textit{g}., }

\newcommand{\splitcell}[1]{\begin{tabular}{@{}c@{}}#1\end{tabular}}

\definecolor{blue-violet}{rgb}{0.54, 0.17, 0.89}

\title{Towards Oracle Knowledge Distillation with Neural Architecture Search}
\author{
	  Minsoo Kang$\textsuperscript{\rm 1,3}$\thanks{Both authors contributed equally} \hspace{0.5cm} Jonghwan Mun$\textsuperscript{\rm 2,3}$\footnotemark[1] \hspace{0.5cm} Bohyung Han$\textsuperscript{\rm 1,3}$ \\
	  $\textsuperscript{\rm 1}$Computer Vision Lab., ASRI, Seoul National University, Korea \\
	  $\textsuperscript{\rm 2}$Computer Vision Lab., POSTECH, Korea \\
	  $\textsuperscript{\rm 3}$Neural Processing Research Center (NPRC), Seoul National University, Korea \\
	  $\textsuperscript{\rm 1}${\{kminsoo,bhhan\}@snu.ac.kr} \hspace{0.3cm}
	  $\textsuperscript{\rm 2}${jonghwan.mun@postech.ac.kr} 
} % All authors must be in the same font size and format. Use \Large and \textbf to achieve this result when breaking a line
\begin{document}
\maketitle

%-------------------------------------------------------------------
%		Abstract
%-------------------------------------------------------------------
\begin{abstract}
We present a novel framework of knowledge distillation that is capable of learning powerful and efficient student models from ensemble teacher networks.
Our approach addresses the inherent model capacity issue between teacher and student and aims to maximize benefit from teacher models during distillation by reducing their capacity gap.
Specifically, we employ a neural architecture search technique to augment useful structures and operations, where the searched network is appropriate for knowledge distillation towards student models and free from sacrificing its performance by fixing the network capacity.
We also introduce an oracle knowledge distillation loss to facilitate model search and distillation using an ensemble-based teacher model, where a student network is learned to imitate oracle performance of the teacher.
We perform extensive experiments on the image classification datasets---CIFAR-100 and TinyImageNet---using various networks.
We also show that searching for a new student model is effective in both accuracy and memory size and that the searched models often outperform their teacher models thanks to neural architecture search with oracle knowledge distillation.
\end{abstract}

%-------------------------------------------------------------------
%		Introduction
%-------------------------------------------------------------------
\section{Introduction}
\label{sec:introduction}

%--------------------------------------------------------------------------
Knowledge Distillation (KD) aims to transfer representations from one model to another, where the source plays a role as a \emph{teacher} while the target becomes a \emph{student} mimicking the representations of the teacher.
KD is widely used to learn a compact student model with the help of a powerful teacher model, \eg a very deep neural network or an ensemble of multiple neural networks.
Existing algorithms related to KD~\cite{BSS,KD,RKD,FitNet,AT,DML} are typically interested in how to improve accuracy by designing an effective training procedure.

%--------------------------------------------------------------------------
Model ensemble~\cite{bagging,boosting,mcl,cmcl,smcl,IE,mclkd} is a useful technique to boost performance using multiple models trained independently (or sometimes jointly).
When we consider Independent Ensemble (IE) in deep neural networks, multiple neural networks of an identical architecture are trained with different random seeds and the final output is determined by model averaging or majority voting.
Although ensemble modeling is effective to achieve high accuracy with moderate amount of effort, its inferences based on simple model averaging or majority voting still have substantial gaps with oracle predictions achieved by the best model selection in the ensemble, and its applicability to resource-hungry system is limited due to the large model size and heavy power consumption.

%--------------------------------------------------------------------------
Although KD with ensemble learning addresses the aforementioned issues partly by transferring the information in an ensemble model to a single network, it is still challenging to train the competitive student compared to the ensemble teacher with a large number of networks. 
Table~\ref{tab:comp_ens} presents our observation about the performance of KD with respect to the number of models for ensemble.
The accuracy of teacher and student improves gradually in general as the number of models increases while students mostly fail to reach accuracy of teachers and its differences are getting larger.
This is partly because a large gap in model capacity between student and teacher hinders learning process of KD as discussed in \cite{TAKD2019}, and the simple objective function to fit the representations of the teacher given by model averaging is not effective to take full advantage of teacher models.
In other words, the limited capacity in the student network becomes a bottleneck of KD, which implies that increasing capacity of student models would be beneficial to reduce the performance gap between teacher and student.
%
%--------------------------------------------------------------------------
\begin{table*}[t!]
	\center
	\caption{
		The performance comparison while varying the number of networks in an ensemble-based teacher model on the CIFAR-100 dataset using ResNet-32 and DenseNet-40-12 networks.
		When the number of ensemble is 1, student and teacher networks are identical.
	}
	\resizebox{0.6\textwidth}{!}
	{
		\begin{tabular}{c|ccc|ccc}
			\toprule		
			\multirow{2}{*}{number of ensemble} & \multicolumn{3}{c|}{ResNet-32} & \multicolumn{3}{c}{DenseNet-40-12} \\
			& Teacher & Student & T-S & Teacher & Student & T-S \\
			\hline \hline
			1  & 69.11 &   -   &   -   & 74.30 &  -    &   -   \\
			2  & 73.77 & 73.84 & -0.07 & 77.47 & 77.82 & -0.35 \\
			3  & 75.57 & 74.12 & 1.45  & 78.70 & 78.03 & 0.67  \\ 
			4  & 76.36 & 74.10 & 2.26  & 79.32 & 78.16 & 1.16  \\
			5  & 76.87 & 74.67 & 2.20  & 79.77 & 78.43 & 1.34  \\
			\bottomrule
		\end{tabular}
	}
	\label{tab:comp_ens}
\end{table*}

%--------------------------------------------------------------------------
We propose an advanced framework for knowledge distillation from ensemble models, which aims to maximize accuracy and efficiency of student networks at the same time.
Contrary to the existing methods assuming that a student model is fixed, we adapt its structure and size, and make it more appropriate for knowledge distillation by alleviating the model capacity gap issue.
We also address how to effectively extract information from a teacher especially when the teacher is an ensemble of multiple models.
To this end, students are made to learn the most accurate information from teacher, which is realized by transferring knowledge from the optimal ensemble combination for each example.
We call this strategy Oracle knowledge Distillation (OD), which encourages student models to achieve oracle accuracy of ensemble teacher models.
Since the inherent model capacity gap incurs critical challenges in KD and OD makes the capacity gap larger, we incorporate neural architecture search with oracle knowledge distillation;
this strategy facilitates to identify the optimal student model with sufficient capacity, which is well-suited for distillation.
In practice, our algorithm searches for a slightly larger model than the backbone student network for effective knowledge distillation, reduces the model capacity gap between student and teacher, and achieves competitive accuracy of the student model.

%--------------------------------------------------------------------------
The main contributions of our work are summarized as follows:
\begin{itemize}[label=$\bullet$]
	\item 
	We propose a novel framework for knowledge distillation by incorporating neural architecture search.  The proposed algorithm addresses capacity issue in KD and aims to identify the optimal structures and operations with adaptive model sizes.
	\item
	Our algorithm introduces a novel oracle knowledge distillation loss, which is particularly useful for an ensemble teacher model.
	We claim that the student networks mimicking oracle predictions have a potential for achieving higher accuracy than the teacher especially when combined with neural architecture search.
	\item
	We demonstrate outstanding performance of the proposed method in diverse settings.  We also make a comprehensive analysis about knowledge distillation from ensemble teacher models, including various issues related to model capacity gap, objective function for architecture search, and loss function for knowledge distillation. 
\end{itemize}

The rest of the paper is organized as follows.
We first discuss several related works.
Then, we describe the details of the proposed framework.
Finally, we present extensive experimental results and concludes our paper.

%-------------------------------------------------------------------
%		Related Work
%-------------------------------------------------------------------
\section{Related Works}
\label{sec:related}

%%%% KD related works
KD is originally proposed to learn compact and fast models and has widely been applied to many practical applications including object detection~\cite{Chen2017LearningEO,Li2017Mimicking}, face recognition~\cite{facemodel} and image retrieval~\cite{chen2018darkrank}.
The main idea of KD is to transfer information from one model to another, and it is realized by learning a student network to mimic the output distributions of a teacher network~\cite{KD}.
Recently, several approaches~\cite{BSS,RKD,FitNet,AT} have been proposed to improve performance of KD.
They address how to extract information better from teacher networks and deliver it to students using the activations of intermediate layers~\cite{FitNet}, attention maps~\cite{AT}, encoded transportable factors~\cite{FT}, decision boundary information obtained by adversarial samples~\cite{BSS} or relational information between training examples~\cite{RKD}.
Also, instead of transferring information from teacher to student in one direction, \cite{DML} proposes a mutual learning strategy, where both models are trained jointly through a bidirectional interactions.

%%%% extensions of KD for other tasks
In addition to model compression by learning small and efficient models, the concept of knowledge distillation is often used for other purposes~\cite{net2net,li2016learning,mclkd,noroozi2018boosting}.
For example, \cite{net2net} proposes a framework incrementally learning larger networks from small networks using knowledge distillation while \cite{li2016learning} employs the idea to overcome a catastrophic forgetting issue in continual learning scenarios.
MCL-KD~\cite{mclkd} utilizes distillation to balance between model specialization and generalization within a multiple choice learning framework for visual question answering.
On the other hand, \cite{noroozi2018boosting} proposes a novel framework of self-supervised learning by distilling learned representations rather than fine-tuning the learned parameters via self supervision for the target tasks.

%--------------------------------------------------------------------------
\begin{figure*}[t]
	\centering
	\includegraphics[width=0.95\textwidth]{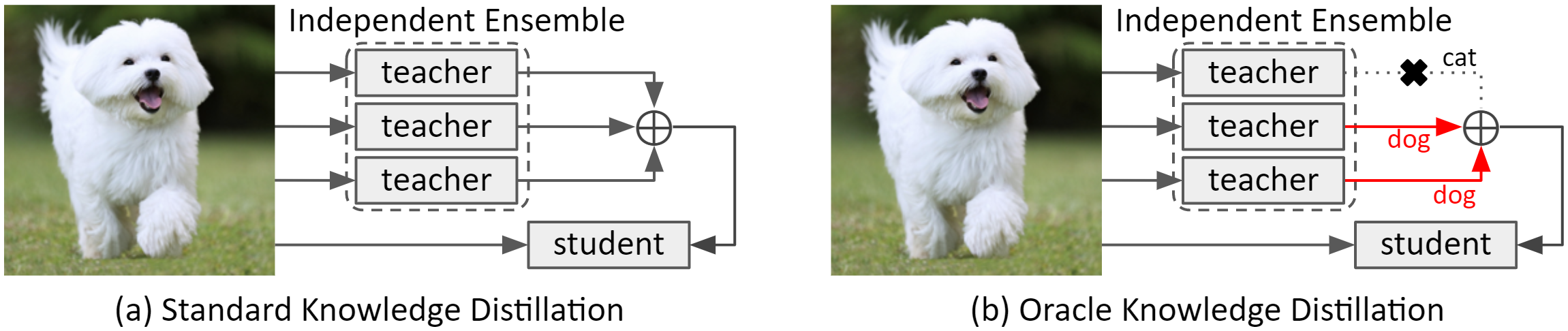}
	\caption{%\small
		Comparison between standard KD and our proposed OD for the ensemble-based teacher model.
		In our approach, we train a student network from only the correct models (red arrows) to imitate the oracle predictions of ensemble teacher.
	}
	\label{fig:comp_KD_OD}
\end{figure*}
%--------------------------------------------------------------------------

%%%% Difference compared to prior works
Our work is different in the following two aspects compared to the existing approaches.
First, we address model capacity issue in a student model by searching for an augmented architecture appropriate for KD while prior methods are interested in how to improve the KD given a student model with a fixed capacity.
Second, although model ensemble is suitable for KD and effectively constructs a powerful teacher model, existing methods rely on a na\"ive knowledge transfer from teachers to students via model averaging or majority voting.
On the contrary, we employ oracle knowledge distillation, which is based on an intuitive and effective loss function for sophisticated knowledge transfer. 
Note that the proposed loss function gives a chance to outperform the teachers with simple model averaging for students.

%%%% Nas related works
Neural Architecture Search (NAS) is an AutoML technique to identify deep neural network models automatically, which is useful to reduce human effort on manual architecture design.
For the purpose, Zoph and Le~\cite{nasnet} employ a RNN controller that searches for the optimal models and is trained to maximize the expected reward---accuracy in validation set---using REINFORCE~\cite{reinforce}.
While NAS often suffers from huge time complexity for model search, ENAS~\cite{enas} accelerates searching process by sharing the weights in the building blocks of all candidate networks.
In addition to the accuracy of target models, MnasNet~\cite{mnasnet} considers inference latency in searching for the optimal model and performs a joint optimization of accuracy and execution time on mobile devices via reinforcement learning.
DARTS~\cite{darts} and NAO~\cite{nao} are gradient-based algorithms realized by a continuous relaxation of architecture representation.
PNAS~\cite{pnas} incorporates a progressive search strategy to reduce search cost using sequential model-based opimization.
Recently, the effectiveness of NAS on image classification tasks leads to a variety of applications to semantic segmentation~\cite{chen2018searching,autodeeplab} and object detection~\cite{detnas}.

From the NAS perspective, our algorithm is unique because it deals with model capacity problems in the context of knowledge distillation.
Note that, instead of searching for a student network architecture from scratch, we start model search from the student network and increase the size of the network to identify the optimal architecture.

%-------------------------------------------------------------------
%		Framework
%-------------------------------------------------------------------
\section{Neural Architecture Search with \\ Oracle Knowledge Distillation}
\label{sec:kdas}
We propose a novel knowledge distillation framework using neural architecture search with an oracle knowledge distillation loss, which is designed for transfer learning from ensemble-based teacher models.
This is motivated by the fact that knowledge distillation is less effective when the capacity gap (\eg the number of parameters) between teacher and student is large as discussed in \cite{TAKD2019}.
Incorporating oracle knowledge distillation loss aggravates the situation by forcing the student to face more challenging task (\ie oracle performance of ensemble model).
Note that since our goal is to address this capacity issue, we are interested in increasing model size from the student network contrary to the more common direction of model compression~\cite{N2N}.
We empirically show that the combination of neural architecture search and oracle knowledge distillation in searching and training networks is particularly helpful to improve both accuracy and efficiency in various scenarios.

%-------------------------------------------------------------
\subsection{Knowledge Distillation}
\label{sec:KD}

KD~\cite{KD} aims to transfer knowledge of a teacher network to a student.
This objective is typically achieved by minimizing the distance between the output distributions of student and teacher.
In other words, given a ground-truth label $y^{(i)}$ and a representation (logit) of a teacher network $l_t^{(i)}$ for an example $x^{(i)}$, a student network learns a representation $l_s^{(i)}$ that minimizes distillation loss $\mathcal{L}_{\text{KD}}$, which is given by
\begin{align}
\mathcal{L}_{\text{KD}} & (l_s^{(i)}, l_t^{(i)}, y^{(i)}) = \\
&\lambda \mathcal{L}_{\text{CE}}(l_s^{(i)}, y^{(i)})  
+ (1-\lambda) \mathcal{L}_{\text{KL}}(l_s^{(i)}, l_t^{(i)}), \nonumber
\label{eq:l_kd}
\end{align}
where $\mathcal{L}_{\text{CE}}(l_s^{(i)}, y^{(i)})$ means Cross-Entropy (CE) loss and $\mathcal{L}_{\text{KL}}(l_s^{(i)}, l_t^{(i)})$ denotes a loss term related to Kullback-Leibler (KL) divergence.
Each term is further defined as
\begin{align}
\mathcal{L}_{\text{CE}}(l_s^{(i)}, y^{(i)}) &= \mathcal{H} (\sigma(l_s^{(i)}), y^{(i)}), \\
\mathcal{L}_{\text{KL}}(l_s^{(i)}, l_t^{(i)}) &= T^2 D_{\text{KL}} (\sigma (l_t^{(i)} / T) || \sigma (l_s^{(i)} / T)),
\end{align}
where $\mathcal{H}(\cdot, \cdot)$ and $D_\text{KL} (\cdot, \cdot)$ are the cross-entropy and the KL-divergence functions, respectively while $\sigma(\cdot)$ denotes a softmax function and $T$ is a temperature parameter.
The student network is trained to predict the correct labels by $\mathcal{L}_{\text{CE}}$ and imitates the output distribution of the teacher network by $\mathcal{L}_{\text{KL}}$, where the two losses are balanced by a hyperparameter $\lambda$.
The softmax distributions of student and teacher networks are softened because the logits are scaled with the temperature $T\ge1$ in $\mathcal{L}_{\text{KL}}$.

Conceptually, any network outperforming student networks can be used as a teacher model.
In this work, we consider the ensemble of student networks as the teacher model due to the following two reasons;
1) model ensemble is a straightforward method to improve accuracy of any state-of-the-art network; 
2) oracle prediction is available in the ensemble model, which allows students to learn better representation and achieve higher accuracy.

%--------------------------------------------------------------------------
\begin{figure*}[t]
	\centering
	\includegraphics[width=0.90\textwidth]{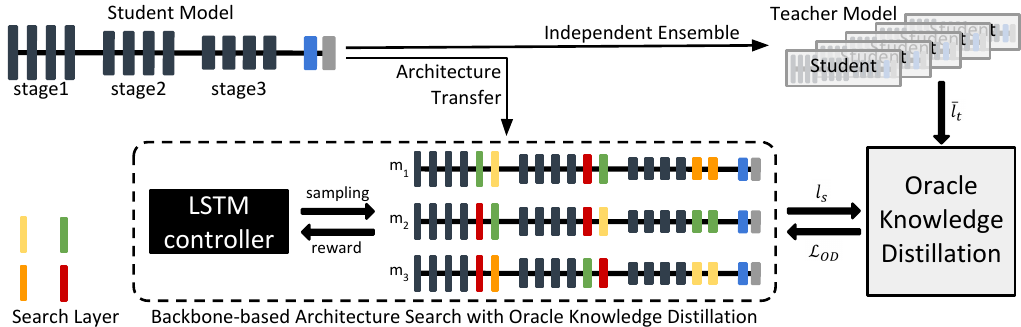}
	\caption{%\small
		%Overview of the proposed knowledge distillation framework using architecture search with the oracle knowledge distillation loss.
		Overview of our algorithm. 
		Given a teacher model based on an ensemble of independently learned multiple student networks, we search for a slightly larger network from a backbone network (\ie student).
		LSTM controller provides candidate networks by sampling add-on operations at the end of individual stages in the student.
		We train the controller by maximizing the expected reward---accuracy of candidates on the validation set---while the candidates are learned with oracle knowledge distillation loss ($\mathcal{L}_{\text{OD}}$). 
		We maintain multiple candidate models throughout the optimization process and the best model below the memory constraint is selected as a new student model and it is re-trained from scratch with $\mathcal{L}_{\text{OD}}$.
		%Given ensemble-based teacher model, we search for an optimal student model with OD, where the candidate of student model is consists of backbone model and searched layers.    
		%1. First of all
		%Note that teacher model is given by the ensemble of independently learned multiple student networks.
	}
	\label{fig:overall_framework}
\end{figure*}
%--------------------------------------------------------------------------
%
\subsection{Oracle Knowledge Distillation}
\label{subsec:od}

Ensemble learning is a powerful technique to improve accuracy by diversifying predictions of multiple models.
In general, on the model distillation, a student network is trained to resemble the average predictions of an ensemble model as illustrated in Figure~\ref{fig:comp_KD_OD}(a) due to the absence of model selection capability;
for an ensemble teacher model with $N$ networks, the teacher logit $l_t^{(i)}$ in Eq.~\eqref{eq:l_kd} is obtained by averaging the logits of the networks, \ie $l_t^{(i)} = \frac{1}{N}\sum_{j=1}^{N} l_{t,j}^{(i)}$.
However, only a subset of the models may predict correct labels, and, consequently, the average predictions may be incorrect;
the ensemble model fails to achieve its oracle accuracy, which can be realized by a proper model selection.

Inspired by this fact, we propose an oracle knowledge distillation loss to improve the performance of distillation from an ensemble teacher.
Let $u_j^{(i)}$ be a binary variable about whether the $j^\text{th}$ model in the ensemble teacher makes a correct prediction for an example $x^{(i)}$.
Then, the student network is trained to minimize the following loss given by the ensemble of $N$ networks:
\begin{equation}
	\begin{aligned}
		\mathcal{L}_{\text{OD}} = 
		\begin{cases}
			\mathcal{L}_{\text{KD}}(l_s^{(i)}, \bar{l}_t^{(i)},y^{(i)})
			& \text{if~}  \sum_{j=1}^{N} u_j^{(i)} > 0 \\
			\mathcal{L}_{\text{CE}}(l_s^{(i)}, y^{(i)})
			& \text{otherwise}
		\end{cases},
	\end{aligned} 
\end{equation}
where
\begin{equation}
\bar{l}_t^{(i)} = \frac{\sum^{N}_{j=1} u_j^{(i)} l^{(i)}_{t,j}}{\sum^{N}_{j=1} u_j^{(i)}}. \nonumber
\end{equation}
$\mathcal{L}_\text{OD}$ encourages the student network to achieve the oracle predictions of the ensemble model by mimicking the average predictions of correct models only as depicted in Figure~\ref{fig:comp_KD_OD}(b).
In the case that there are no correct models, we make the student network fit to ground-truth labels.
Since the accuracy given by oracle predictions is always better than average predictions, the trained student network has a potential to outperform its teacher model that employs model averaging and/or majority voting.

%--------------------------------------------------------------------------
\subsection{Optimal Model Search for Knowledge Distillation}
\label{subsec:bas}
The capacity issue in a student makes the student network fail to take full advantage of ensemble model.
To overcome this challenge, we propose a novel Knowledge Distillation framework with Architecture Search (KDAS) using the proposed oracle knowledge distillation loss designed for ensemble-based teacher models.
In the proposed framework, our goal is to find a slightly larger network with sufficient capacity for distillation than the original student models.
For the purpose, we perform the operation search from the architecture of a student and select the final model under the memory constraint as illustrated in Figure~\ref{fig:overall_framework}.

%%%%%
\subsubsection{Architecture search from backbone models}
Most architecture search algorithms~\cite{pnas,darts,nao,nasnet} search for an optimal architecture from scratch.
However, we perform a backbone-based architecture search, where the architecture of the student network is used as the starting point (\ie backbone) and we augment a set of operations to the backbone model during the search procedure.
This strategy can reduce search space significantly and facilitate to stabilize the training process.

For the efficient and effective neural architecture search to build a larger network from a backbone model, we put add-on operations after the individual stages of the standard convolutional neural networks;
modern convolutional neural networks typically consist of a series of identical components, each of which is called by stage, \eg stacked convolutions in VGGNet~\cite{vggnet} and stacked building blocks in ResNet~\cite{resnet}.
When human manually designs higher capacity of the network, each stage is often made deeper by adding more operations~\cite{resnet,densenet}.
Inspired by this convention, we perform operation search at the end of individual stages in the backbone network and identify competitive model with minimal efforts.
Assuming that the student backbone network $S$ has $k$ stages for feature extraction, \ie $S = \{s_1, s_2, ..., s_k\}$, the searched network $\hat{S}$ is represented by $\hat{S} = \{s_1, \hat{o}_1, s_2, \hat{o}_2, ..., s_k, \hat{o}_k\}$, where $\hat{o} = \{\hat{o}_1, ..., \hat{o}_k\}$ denotes a set of add-on operations.
We do not perform operation search in a classifier consisting of the global pooling and fully-connected layers.

%%%%%
\subsubsection{Search space} 
We search for a network facilitating distillation based on 7 operations with skip connections: an identity operation, convolutions with filter sizes 3$\times$3 and 5$\times$5, depthwise-separable convolutions with filter sizes 3$\times$3 and 5$\times$5, and max pooling and average pooling of kernel size 3$\times$3.
Note that the output channel dimension of each new convolutional layer is set to the input channel dimension of the layer.
Overall, there are $2^{\frac{L(L-1)}{2}} \times 7^{L}$ candidate networks in the whole search space when we add $L$ layers in total to the backbone model.

%%%%%
\subsubsection{Optimization}
We train an LSTM controller to sample an architecture for the oracle knowledge distillation from the predefined search space;
learning the LSTM controller is achieved via REINFORCE~\cite{reinforce} by alternating the following two steps:
1) sampling an architecture using the LSTM controller and training it for the predefined number of iterations with the proposed oracle distillation loss ($\mathcal{L}_{\text{OD}}$) and 2) updating the LSTM controller based on the reward of the trained model.

Specifically, let us denote the learnable parameters of the LSTM controller by $\theta$.
Then, the controller is trained to maximize the expected reward $J(\theta)$ as follows:
\begin{equation}
	J(\theta) \equiv \mathbb{E}_{\mathbf{m} \thicksim \pi(\mathbf{m};\theta)} [\mathit{R}(\mathbf{m})],
\end{equation}
where $\pi(\mathbf{m};\theta)$ is a policy of the controller and $R(\mathbf{m})$ denotes the reward (\ie validation accuracy) from a sampled architecture $\mathbf{m}$.
Using Monte-Carlo methods, the expected gradient on the sampled architectures is given by
\begin{align}
	\nabla_{\theta} J(\theta) &= \mathbb{E}_{\mathbf{m} \thicksim \pi(\mathbf{m};\theta)} [\mathit{R}(\mathbf{m}) \nabla_{\theta} \log\pi(\mathbf{m};\theta)] \nonumber \\
	& \approx \frac{1}{N} \sum_{j=1}^{N} [\mathit{R}(\mathbf{m}_j) \nabla_{\theta} \log\pi(\mathbf{m}_j;\theta)],
	\label{eq:monte}
\end{align}
where $N$ is the number of sampled architectures.
To reduce the variance, the reward $\mathit{R}(\mathbf{m}_j)$ is replaced by $\mathit{R}(\mathbf{m}_j)-b$, where $b$ is a baseline function given by a moving average of the past rewards.
Following ENAS~\cite{enas}, we perform an efficient learning scheme; certain layers with their parameters in multiple candidate networks can be shared and the reward is computed only on a batch rather than a whole validation set.

%%%%%
\subsubsection{Network selection with memory constraint}
After convergence of the controller, we select the most accurate and efficient model as follows:
\begin{equation}
\mathbf{m}^{*} = \argmax_{m} R(\mathbf{m}),  
~~~~\text{s.t.}~~ |\mathbf{m}| \leq M,
\end{equation}
where $|\mathbf{m}|$ denotes the amount of memory spent to store the parameters in a sampled network $\mathbf{m}$ and $M$ is the memory constraint for model selection.
The selected model is re-trained from scratch with the proposed oracle knowledge distillation loss $\mathcal{L}_{\text{OD}}$.

%-------------------------------------------------------------------
%		Experiments
%-------------------------------------------------------------------
\section{Experiments}
\label{sec:experiments}
This section presents performance of the proposed algorithm in comparison to existing methods.
We also discuss characteristics of our approach obtained from in-depth analysis.

%--------------------------------------------------------------------------
\begin{table*}[t!]
	\centering
	\caption{
		Ablation studies varying combinations of loss in searching ($\mathcal{L}_{\text{S}}$) and training ($\mathcal{L}_{\text{T}}$) networks on CIFAR-100 and TinyImageNet datasets.
		We employ a ResNet-32 network as the student model and ensemble of five student networks (\ie ResNet-32$\times$5) as the teacher model whose oracle accuracies ($\%$) are 87.79 and 75.30 in CIFAR-100 and TinyImageNet.
		The bold-faced and red-colored numbers mean the best algorithm in accuracy for each model and each dataset, respectively.
	}
	\resizebox{0.8\textwidth}{!}{
		\begin{tabular}{c|c|cc|cc|cc|c}
			\toprule
			& \multirow{2}{*}{Model} & \multirow{2}{*}{$\mathcal{L}_{\text{S}}$} & \multirow{2}{*}{$\mathcal{L}_{\text{T}}$} 
			& \multicolumn{2}{c|}{CIFAR-100} & \multicolumn{2}{c|}{TinyImageNet} & \multirow{2}{*}{\splitcell{Network\\identified by}}
			\\
			&  &  &  & \multicolumn{1}{c}{Accuracy} & Memory & \multicolumn{1}{c}{Accuracy} & Memory &
			\\
			\hline \hline
			
			% teacher model
			\multicolumn{1}{c|}{M1}  & Teacher     & - & $\mathcal{L}_{\text{CE}}$  & \multicolumn{1}{c}{76.87} & 2.35M & \multicolumn{1}{c}{62.59} & 2.38M & \multirow{4}{*}{-}
			\\
			\cline{1-8}
			
			% student model
			\multicolumn{1}{c|}{M2}  & \multirow{3}{*}{Student} & \multirow{3}{*}{-} & $\mathcal{L}_{\text{CE}}$ & \multicolumn{1}{c}{69.11 $\pm$ 0.24} & \multirow{3}{*}{0.47M} & \multicolumn{1}{c}{54.14 $\pm$ 0.65} & \multirow{3}{*}{0.48M} & 
			\\
			\multicolumn{1}{c|}{M3}  & & & $\mathcal{L}_{\text{KD}}$ & \multicolumn{1}{c}{74.67 $\pm$ 0.10} & & \multicolumn{1}{c}{\textbf{58.68 $\pm$ 0.09}} & &
			\\
			\multicolumn{1}{c|}{M4}  & & & $\mathcal{L}_{\text{OD}}$ & \multicolumn{1}{c}{\textbf{74.77 $\pm$ 0.02}} & & \multicolumn{1}{c}{58.66 $\pm$ 0.25} & &
			\\
			\hhline{=========}
			
			%% MMN of ResNet-62
			\multicolumn{1}{c|}{M5}  & \multirow{3}{*}{ResNet-62}   & \multirow{3}{*}{-} & $\mathcal{L}_{\text{CE}}$ & \multicolumn{1}{c}{   72.06 $\pm$ 0.31} & \multirow{3}{*}{0.96M} & \multicolumn{1}{c}{58.62 $\pm$ 0.16} & \multirow{3}{*}{0.97M} & \multirow{6}{*}{Man-Made}
			\\ 
			\multicolumn{1}{c|}{M6}  & & & $\mathcal{L}_{\text{KD}}$ & \multicolumn{1}{c}{\textbf{76.09 $\pm$ 0.20}} & & \multicolumn{1}{c}{61.05 $\pm$ 0.31} & &
			\\
			\multicolumn{1}{c|}{M7}  & & & $\mathcal{L}_{\text{OD}}$ & \multicolumn{1}{c}{75.89 $\pm$ 0.19} & & \multicolumn{1}{c}{\textbf{61.25 $\pm$ 0.14}} & &
			\\
			\cline{1-8}
			
			%% MMN of ResNet-110			
			\multicolumn{1}{c|}{M8}  & \multirow{3}{*}{ResNet-110}  & \multirow{3}{*}{-} & $\mathcal{L}_{\text{CE}}$ & \multicolumn{1}{c}{73.77 $\pm$ 0.19} & \multirow{3}{*}{1.73M} & \multicolumn{1}{c}{60.24 $\pm$ 0.45} & \multirow{3}{*}{1.74M} &
			\\
			\multicolumn{1}{c|}{M9}  & & & $\mathcal{L}_{\text{KD}}$ & \multicolumn{1}{c}{\textbf{76.77 $\pm$ 0.52}} & & \multicolumn{1}{c}{62.03 $\pm$ 0.03} & &
			\\ 
			\multicolumn{1}{c|}{M10} & & & $\mathcal{L}_{\text{OD}}$ & \multicolumn{1}{c}{76.68 $\pm$ 0.17} & & \multicolumn{1}{c}{\textbf{62.66 $\pm$ 0.53}} & &
			\\
			\hline\hline
			
			% student and teacher with $\mathcal{L}_{\text{CE}}$< $\mathcal{L}_{\text{KD}}$ and $\mathcal{L}_{\text{OD}}$ w/ NAS
			\multicolumn{1}{c|}{M11} & \multirow{3}{*}{NAS} & \multirow{3}{*}{$\mathcal{L}_{\text{CE}}$} & $\mathcal{L}_{\text{CE}}$ 
			& \multicolumn{1}{c}{74.55 $\pm$ 0.51} & \multirow{3}{*}{0.97M} & \multicolumn{1}{c}{62.01 $\pm$ 0.60} & \multirow{3}{*}{0.90M} & \multirow{9}{*}{AutoML}
			\\
			\multicolumn{1}{c|}{M12} & & & $\mathcal{L}_{\text{KD}}$ 
			& \multicolumn{1}{c}{76.85 $\pm$ 0.33} & & \multicolumn{1}{c}{62.10 $\pm$ 0.17} & &
			\\
			\multicolumn{1}{c|}{M13} & & & $\mathcal{L}_{\text{OD}}$ 
			& \multicolumn{1}{c}{\textbf{77.05 $\pm$ 0.23}} & & \multicolumn{1}{c}{\textbf{62.57 $\pm$ 0.11}} & &
			\\
			\cline{1-8}
			
			\multicolumn{1}{c|}{M14} & \multirow{3}{*}{KDAS (ours)} & \multirow{3}{*}{$\mathcal{L}_{\text{KD}}$} & $\mathcal{L}_{\text{CE}}$ 
			& \multicolumn{1}{c}{74.56 $\pm$ 0.35 } & \multirow{3}{*}{0.93M} & \multicolumn{1}{c}{\textbf{62.92 $\pm$ 0.10}} & \multirow{3}{*}{0.95M} &
			\\
			\multicolumn{1}{c|}{M15} & & & $\mathcal{L}_{\text{KD}}$ 
			& \multicolumn{1}{c}{76.97 $\pm$ 0.08} & & \multicolumn{1}{c}{62.34 $\pm$ 0.10} & &
			\\
			\multicolumn{1}{c|}{M16} & & & $\mathcal{L}_{\text{OD}}$ 
			& \multicolumn{1}{c}{\textbf{77.04 $\pm$ 0.33}} & & \multicolumn{1}{c}{62.73 $\pm$ 0.09} & &
			\\
			\cline{1-8}
			
			\multicolumn{1}{c|}{M17} & \multirow{3}{*}{KDAS (ours)} & \multirow{3}{*}{$\mathcal{L}_{\text{OD}}$} & $\mathcal{L}_{\text{CE}}$ 
			& \multicolumn{1}{c}{75.14 $\pm$ 0.26} & \multirow{3}{*}{0.89M} & \multicolumn{1}{c}{62.60 $\pm$ 0.11} & \multirow{3}{*}{0.87M}  &
			\\
			\multicolumn{1}{c|}{M18} & & & $\mathcal{L}_{\text{KD}}$ 
			& \multicolumn{1}{c}{76.92 $\pm$ 0.33} & & \multicolumn{1}{c}{62.17 $\pm$ 0.12} & &
			\\
			\multicolumn{1}{c|}{M19} & & & $\mathcal{L}_{\text{OD}}$ 
			& \multicolumn{1}{c}{{\color{red}\textbf{77.27 $\pm$ 0.11}}} & & \multicolumn{1}{c}{{\color{red}\textbf{63.04 $\pm$ 0.17}}} & &
			\\
			\bottomrule			
		\end{tabular}
	}
	\label{tab:ablation}
\end{table*}
%--------------------------------------------------------------------------

%--------------------------------------------------------------------------
\subsection{Experimental Setup}
\label{subsec:exp_setup}

%%%%%
\subsubsection{Datasets}
We evaluate our algorithm on the image classification task using CIFAR-100 and TinyImageNet datasets.
CIFAR-100 dataset~\cite{cifar09} is composed of 50,000 training and 10,000 testing images in 100 classes, where the size of image is 32$\times$32.
TinyImageNet dataset contains 100,000 and 10,000 images from 200 object classes with their size 64$\times$64 for training and validation, respectively.
For architecture search, 10\% of training images are held out as training-validation set to compute reward.
For both datasets, we perform the preprocessing of subtracting means and dividing by standard deviations in individual RGB channels, and employ the standard data augmentation techniques such as random cropping with zero padding and horizontal flipping.

%%%%%
\subsubsection{Implementation details}
We employ publicly available ENAS~\cite{enas} code\footnote{https://github.com/melodyguan/enas} for neural architecture search implementation in TensorFlow~\cite{abadi2016tensorflow}. %~\cite{tensorflow2015-whitepaper}.
Given a searched network or networks designed by human, we optimize the networks for 300 epochs using SGD with Nesterov momentum~\cite{nesterov} of 0.9, a weight decay of 0.0001 and a batch size of 128.
Following \cite{lan2018knowledge}, the initial learning rate is set to 0.1, and is divided by 10 at 150$^\mathrm{th}$ and 225$^\mathrm{th}$ epoch, respectively.
We also perform warm-up strategy~\cite{resnet} with learning rate of 0.01 with ResNet-110 until 400$^{\mathrm{th}}$ and 900$^{\mathrm{th}}$ iterations for CIFAR-100 and TinyImagenet datasets, respectively. 
For KD and OD, a temperature $T$ is fixed 3 and a balancing factor $\lambda$ is set to 0.
We train all models 3 times and report their average and standard deviation for score reports. 
%We will release the source code to facilitate understanding and reproduction of our algorithm.

%--------------------------------------------------------------------------
\subsection{Experimental Results}
\label{subsec:results}

%%%%%
\subsubsection{Main results}
We perform extensive experiments on the two datasets to investigate the effectiveness of architecture search and oracle knowledge distillation loss.
In this experiment, we consider the following types of models,
(\romannum{1}) M1: a teacher model of ResNet-32$\times$5,
(\romannum{2}) M2-4: a student model of ResNet-32,
(\romannum{3}) M5-10: two baseline models designed by human, which are ResNet-62 and ResNet-110 having 2x and 4x model size of student,
(\romannum{4}) M11-19: three variants of KDAS models searched by different losses, where we constraint model size up to 2x of student
We train all models except teacher using one out of three loss functions, $\mathcal{L}_{\text{CE}}$, $\mathcal{L}_{\text{KD}}$ and $\mathcal{L}_{\text{OD}}$, as presented in Table~\ref{tab:ablation}.

Table~\ref{tab:ablation} presents the results and we can obtain following observations.
First, as we claimed earlier, the student networks trained by distillation losses (M3-4) suffer from capacity and complexity issues and fail to achieve competitive accuracy.
Second, manually designed networks with higher capacity (\ie ResNet-62 and ResNet-110) partly address the capacity issue, where their accuracy is proportional to the memory size of network.
This result implies that increasing capacity of student networks is helpful for reducing performance gap between student and teacher.
It is also noticeable that ResNet-110 in TinyImageNet can achieve marginally higher accuracy than teacher by applying $\mathcal{L}_{\text{OD}}$.
Third, the networks identified by KDAS with distillation (M15-16, M18-19) are consistently better than Man-Made Networks (MMNs), ResNet-62 and ResNet-110, even with smaller memory size.
Fourth, training networks with $\mathcal{L}_{\text{OD}}$ shows the best accuracy in most cases of KDAS.
One exception is when the model is searched with $\mathcal{L}_\text{KD}$ but trained with $\mathcal{L}_\text{CE}$.
In this case, the network outperforms the teacher even without distillation, which implies that the teacher would have a negative impact on the student with $\mathcal{L}_\text{KD}$.
However, $\mathcal{L}_\text{OD}$ still encourages a student to learn better representation and the student outperforms the teacher model (see M14-16 on TinyImageNet).
Finally, the full KDAS model (M19) with $\mathcal{L}_{\text{OD}}$ in both architecture search and training achieves the best accuracy even including teacher model.
This result shows that the combination of architecture search with $\mathcal{L}_{\text{OD}}$ is particularly well-suited for improving accuracy of a student network.
%
%--------------------------------------------------------------------------
\begin{figure}
	\centering
	\includegraphics[width=0.90\columnwidth]{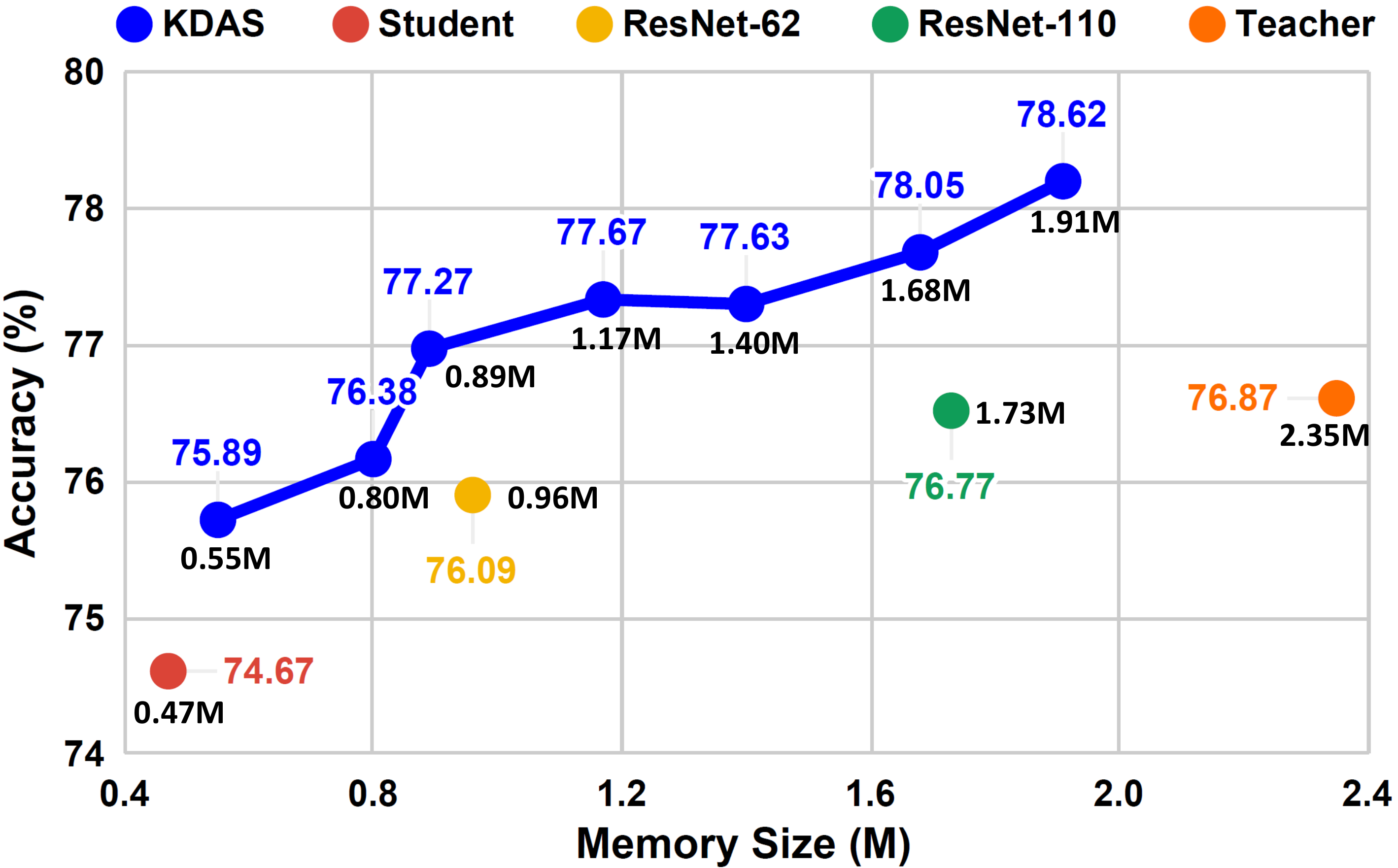}
	\caption{%\small
		Accuracies varying memory size of networks given by KDAS on the CIFAR-100 dataset with the backbone student network ResNet-32.
	} 
	\label{fig:res_varying_mem}
\end{figure}
%--------------------------------------------------------------------------
%
%%%%%

\subsubsection{Comparison with traditional NAS}
To validate the necessity of NAS along with knowledge distillation, we conduct NAS without a distillation loss in searching for networks and then train the searched networks with KD and OD losses.
As presented in Table~\ref{tab:ablation}, the NAS baseline networks (M12-13) perform worse than the architectures identified by KDAS with OD (M18-19).
Also, note that even with $\mathcal{L}_{\text{CE}}$, the network identified by KDAS (M17) provides better performance than the counterpart (M11).

%--------------------------------------------------------------------------
\begin{table*}[t!]
	\center
	\caption{
		Results with various networks on the CIFAR-100 dataset.
		We use ResNet-218, WideResNet-76-1, WideResNet-28-2, WideResNet-28-4 networks as MMN of student ResNet-110, WideResNet-40-1, WideResNet-16-2 networks, and WideResNet-16-4 networks, respectively.
		Numbers in red and blue denote the best and second-best models including the teacher model.
	}
	\resizebox{0.985\textwidth}{!}
	{
		\begin{tabular}{c|cc|cc|cc|cc|cc}
			\toprule		
			\multirow{2}{*}{Method} & \multirow{2}{*}{$\mathcal{L}_{\text{S}}$} & \multirow{2}{*}{$\mathcal{L}_{\text{T}}$} 
			& \multicolumn{2}{c|}{ResNet-110} & \multicolumn{2}{c|}{WideResNet-40-1} & \multicolumn{2}{c|}{WideResNet-16-2} & \multicolumn{2}{c}{WideResNet-16-4} \\
			&                           &                           & Accuracy   & Memory & Accuracy   & Memory & Accuracy   & Memory & Accuracy   & Memory \\
			\hline \hline
			Teacher  & -                         & $\mathcal{L}_{\text{CE}}$ & {\color{red}\textbf{79.24}} & 8.67M & {\color{red}\textbf{77.53}} & 2.85M & {\color{blue}\textbf{77.77}} & 3.52M & {\color{blue}\textbf{79.49}} & 13.86M
			\\
			Student  & -                         & $\mathcal{L}_{\text{CE}}$ & 73.77 $\pm$ 0.19 & 1.73M & 69.96 $\pm$ 0.15 & 0.57M & 71.16 $\pm$ 0.30 & 0.70M & 75.17 $\pm$ 0.24 & 2.77M
			\\
			Student  & -                         & $\mathcal{L}_{\text{KD}}$ & 76.77 $\pm$ 0.52 & 1.73M & 74.72 $\pm$ 0.23 & 0.57M & 75.42 $\pm$ 0.04 & 0.70M & 78.59 $\pm$ 0.34 & 2.77M
			\\%		\hline
			MMN      & -                         & $\mathcal{L}_{\text{KD}}$ & 77.39 $\pm$ 0.21 & 3.48M &76.48 $\pm$ 0.15 & 1.15M & 76.97 $\pm$ 0.05 & 1.48M & 79.28 $\pm$ 0.16 & 5.87M
			\\		\hline
			KDAS     & $\mathcal{L}_{\text{OD}}$ & $\mathcal{L}_{\text{OD}}$ & {\color{blue}\textbf{79.01 $\pm$ 0.28}} & 2.73M & {\color{blue}\textbf{76.70 $\pm$ 0.25}} & 1.14M & {\color{red}\textbf{77.83 $\pm$ 0.23}} & 1.30M & {\color{red}\textbf{79.79 $\pm$ 0.24}} & 5.47M \\
			\bottomrule
		\end{tabular}
	}
	\label{tab:comp_various_net}
\end{table*}

%--------------------------------------------------------------------------
\begin{table*}[t!]
	\center
	\caption{
		Performance comparison with other KD algorithms on the CIFAR-100 dataset.
		We use a single ResNet-110 network as a teacher model.
		The red-colored number means the highest accuracy.
	}
	\resizebox{0.89\textwidth}{!}
	{
		\begin{tabular}{c|c|c|c|c|c||c}
			\toprule		
			Student & CE & KD & DML & BSS & TAKD & KDAS (0.91M) 
			\\
			\hline \hline
			ResNet-62 (0.96M) & 71.73 $\pm$ 0.03 & 74.57 $\pm$ 0.18 & 72.98 $\pm$ 1.07 & 73.06 $\pm$ 0.53 & 75.18 $\pm$ 0.13 & \multirow{2}{*}{\color{red}\textbf{75.82 $\pm$ 0.32}}
			\\
			ResNet-68 (1.05M) & 71.77 $\pm$ 0.06 & 74.82 $\pm$ 0.09 & 73.39 $\pm$ 0.70 & 73.43 $\pm$ 0.21 & 75.45 $\pm$ 0.12 & 
			\\
			\bottomrule
		\end{tabular}
	}
	\label{tab:comp_other_KD}
\end{table*}
%--------------------------------------------------------------------------

%%%%%
\subsubsection{Analysis on varying memory size}
We analyze performance of student models with different sizes obtained by KDAS on CIFAR-100 dataset.
Note that teacher is based on five ResNet-32 models.
We search for the student network architectures by varying memory constraint from $1.5\times$ to $4{\times}$ of the base student model (0.47M).
For comparison, we also train MMNs using the standard knowledge distillation loss ($\mathcal{L}_{\text{KD}}$), which are selected to be $2{\times}$ larger than the base student model.
Figure~\ref{fig:res_varying_mem} shows that KDAS models outperform MMNs in terms of both accuracy and efficiency; these results support the benefit of architecture search for KD.
Note that KDAS models larger than 0.89M ($2\times$ larger than the base model) achieve better accuracies than the teacher model thanks to the oracle knowledge distillation loss.

%%%%%
\subsubsection{Evaluation with other networks}
To validate the model-agnostic property of our algorithm, various student models have been evaluated on CIFAR-100 dataset.
Table~\ref{tab:comp_various_net} summarizes the results from the tested backbone architectures including ResNet-110, WideResNet-40-1, WideResNet-16-2, and WideResNet-16-4~\cite{WRN}, where networks identified by KDAS with OD consistently outperform corresponding MMNs with comparable memory sizes while showing competitive or outperforming performance compared to teacher models.

%%%%%
\subsubsection{Comparison with other KD methods}
Although our work focuses on distillation from ensemble teacher models, we also compare KDAS with other KD approaches including KD~\cite{KD}, DML~\cite{DML}, BSS~\cite{BSS} and TAKD~\cite{TAKD2019}.
For the fair comparisons with other KD methods, we follow  BSS~\cite{BSS} implementation environments\footnote{https://github.com/bhheo/BSS\_distillation} using PyTorch~\cite{pytorch}.
Contrary to the main experiments, we use a single model teacher given by ResNet-110 and train two student networks of ResNet-62 and ResNet-68 for other KD methods while we search for a student network with comparable size to ResNet-62 from a base model of ResNet-32 and train the searched network with KD.
For TAKD, we employ ResNet-86 as a teacher assistant (TA) network.
Table~\ref{tab:comp_other_KD} shows that the model identified by KDAS outperforms other methods in both accuracy and memory size even without the proposed oracle knowledge distillation loss, which is not available in this experiment.
This result implies that searching for student architectures is a promising direction for knowledge distillation.

%--------------------------------------------------------------------------
\begin{table}[t!]
	\centering
	\caption{
		Training accuracy of single ResNet-32 network on CIFAR-100 and TinyImageNet datasets. 
		We also present the percentage of training examples in terms of the number of models that predict correctly.
		%the meaning of numbers denote to the percentage of training examples for the number of models that predict the correct answer.
	}
	\resizebox{0.995\columnwidth}{!}
	{
		\begin{tabular}{c|ccccc|c}
			\toprule
			\multirow{2}{*}{Dataset}  & \multicolumn{5}{c|}{\# of models that predict correctly}  & \multirow{2}{*}{\splitcell{Training\\Acc.}} \\
			&  \multicolumn{1}{c}{1}  & \multicolumn{1}{c}{2}  & \multicolumn{1}{c}{3}  & \multicolumn{1}{c}{4}  & \multicolumn{1}{c|}{5} &  \\ 
			\hline\hline
			CIFAR-100     &  \multicolumn{1}{c}{0.5}  &  \multicolumn{1}{c}{1.0}  &  \multicolumn{1}{c}{2.6}  &  \multicolumn{1}{c}{8.9}  &  \multicolumn{1}{c|}{86.9}  &  94.04  \\
			TinyImageNet  &  \multicolumn{1}{c}{6.3}  &  \multicolumn{1}{c}{6.8}  &  \multicolumn{1}{c}{8.7}  &  \multicolumn{1}{c}{14.3}  &  \multicolumn{1}{c|}{49.6}  &  70.28 \\
			\bottomrule
		\end{tabular}
	}
	\label{tab:stat}
\end{table}
%---------------------------------------------------------------------------

\subsubsection{Practical benefit of OD over KD}  
The advantage of OD over KD comes from their different characteristics on training examples.
OD may be useless if all the constituent models in an ensemble teacher network predict correct answers for most of training examples.
However, in reality, the average training accuracy of all 5 models in the teacher is clearly lower than 100\% as seen in Table~\ref{tab:stat}. 
We also present the distribution of the training examples in terms of the number of models predicting correct answers, and the statistics demonstrates that many training examples in each dataset can take advantage of OD.
In particular, the objectives of OD and KD are different for more than half of the training examples (\ie 50.4\%) in TinyImageNet, and it implies that the accuracy gains in OD can be significant compared to KD.
Indeed, Table~\ref{tab:ablation} illustrates that the benefit of OD is more salient in TinyImageNet, which is natural because the dataset is more challenging and the trained models are less competitive.

%-------------------------------------------------------------------
%		Conclusion
%-------------------------------------------------------------------
\section{Conclusion}
\label{sec:conclusion}
We propose a novel framework of oracle knowledge distillation with neural architecture search especially designed for ensemble teacher models.
The proposed framework addresses the capacity and complexity issues, and aims to find a desirable architecture with additional memory, which mimics oracle predictions of ensemble model more accurately.
We empirically show that the combination of architecture search and oracle knowledge distillation successfully provides high-performance student models in both accuracy and memory size, and most of the searched networks achieve competitive performances to teacher models.
We believe that our framework searching for an optimal student network is a promising research direction of knowledge distillation.

{
\small
\paragraph{Acknowledgments}
We truly thank Tackgeun You for helpful discussion. This work was partly supported by Samsung Advanced Institute of Technology (SAIT) and Institute for Information \& Communications Technology Promotion (IITP) grant funded by the Korea government (MSIT) [2017-0-01778, 2017-0-01780].
}
% Korean ICT R\&D program of the MSIP/IITP grant

{
\small
\bibliography{references}
\bibliographystyle{aaai}
}

\end{document}